\documentclass[manuscript,screen]{acmart}
\AtBeginDocument{%
  \providecommand\BibTeX{{%
    \normalfont B\kern-0.5em{\scshape i\kern-0.25em b}\kern-0.8em\TeX}}}

\usepackage{textgreek}
\usepackage[utf8]{inputenc}

\usepackage[T1]{fontenc}
\usepackage[greek,spanish]{babel}
\usepackage{tipa}

\usepackage{newunicodechar}
\usepackage{textcomp}
\usepackage{amsmath}
\setcopyright{acmcopyright}
\copyrightyear{2018}
\acmYear{2018}
\acmDOI{10.1145/1122445.1122456}

\begin{document}

\title{Perception Point: Identifying Critical Learning Periods in Speech for Bilingual Networks}

\author{Anuj Saraswat}
\authornote{All three authors contributed equally to this research.}
\email{anuj.saraswat@midas.center}
\affiliation{%
  \institution{IIIT Delhi}
  \country{India}
}
\author{Mehar Bhatia*}
\affiliation{%
  \institution{IIIT Delhi}
  \country{India}}
\email{mehar.bhatia@midas.center}

\author{Yaman Kumar Singla*}
\affiliation{%
  \institution{IIIT Delhi}
  \country{India}
}
\email{yamank@iiitd.ac.in}

\affiliation{%
  \institution{Adobe}
  \country{India}
}
\affiliation{%
  \institution{State University of New York at Buffalo}
  \country{USA}
}

\author{Changyou Chen}
\affiliation{%
 \institution{State University of New York at Buffalo}
 \country{USA}}

\author{Rajiv Ratn Shah}
\affiliation{%
  \institution{IIIT Delhi}
  \country{India}}
\email{rajivratn@iiitd.ac.in}

\begin{abstract}
  Recent studies in speech perception have been closely linked to fields to cognitive psychology, phonology, and phonetics in linguistics. During perceptual attunement, a critical and sensitive developmental trajectory has been examined in bilingual and monolingual infants where they can best discriminate common phonemes. In this paper, we compare and identify these cognitive aspects on deep neural-based visual lip-reading models. We conduct experiments on the two most extensive public visual speech recognition datasets for English and Mandarin. Through our experimental results, we observe a strong correlation between these theories in cognitive psychology and our unique  modeling. We inspect how these computational models develop similar phases in speech perception and acquisitions.
\end{abstract}


\maketitle

\section{Introduction}
Speech Perception has been seen as the process by which the sounds of a language are heard, seen, interpreted and understood. Since several decades, researchers have been looking into various aspects of speech perception, from finding perceptual or acoustic cues in speech signals, to highlighting the fundamental units and theories. Researchers in the field of experimental and developmental psychology have also shown deep interest in area of speech perception, studying whether speech perception is rather an active sequential or categorical cognitive process. Many perceptual learning theories have also combined with findings of language acquisition, phonological development and word learning. While focusing on linguistic effects on speech perception, it has been observed that early language experience profoundly affects how speech sounds are perceived throughout life. The influence of linguistic experience begins very early in development \cite{kuhl2004early}. A rich literature has documented that by 12 months of age, infants’ ability to discriminate nonnative consonants already begins to decline while the ability to discriminate native consonants improves \cite{kuhl2008phonetic, kuhl2006infants, werker1984cross}. For adults, discrimination of non-native consonants remains far from `native-like', even after periods of intensive training \cite{logan1991training, flege1999age, lim2011learning}.

Children acquire language mainly through listening, watching, and imitating language before they know any words. By the time they are born, it has been well researched that human infants are already well prepared for language acquisition and show a preference for speech over other types of complex sound \cite{vouloumanos2007listening}. During the period of perceptual attunement, there comes a stage where the bilingual infants can discriminate best between phoneme distinctions which are most common across the two languages and collapse the less frequent ones; we term this period as the \textsc{critical period} \cite{bosch2003simultaneous, sundara2008development}. The same developmental trajectory is observed in bilingual and monolingual infants if more sensitive testing paradigms are used \cite{albareda2011acquisition}. 
Our main contributions in this work are summarized: 
\begin{itemize}
    \item We perform experiments on state-of-the-art visual lip-reading models in a bilingual and monolingual setting and determine the occurrence of a critical and sensitive period.
    \item We study how critical periods effect language acquisition and derive strong correlations with theories in cognitive science, psychology and linguistics with deep neural based models.
\end{itemize}

\section{Methodology}

Visual speech recognition (VSR) combines deep learning approaches and large-scale datasets with audio and videos. Recent advances in visual speech recognition have achieved remarkable recognition rates, even surpassing human performance. However, interpreting how deep neural networks acquire language is still an under-studied area. We try to explore this problem and draw parallels with how humans learn languages by conducting experiments in monolingual and bilingual training environments. 

An early and essential task for humans is to make sense of the speech, requiring attention to audio and visual cues. We focus on the visual part of the language learning using a state-of-the-art visual speech recognition model described in \cite{petridis2018audio}. Visual speech recognition models learn to recognize speech from lip movements and other facial cues. VSR models try to predict visemes which are the visual equivalent of a phoneme. 

Literature from cognitive science and linguistics states the occurrence of a sensitive or critical period during the phase of language acquisition during which infants can best discriminate the phoneme distinctions that are most common across the two languages. Psycho-acoustic tests have indicated that human vision classifies visemes (visual representation of phonemes) into different classes \cite{alothmany2010classification}. We convert phonemes to visemes, details of which are listed in the experiment section. Moreover, we try to observe occurrence of a critical period in visual lip reading model and draw comparisons between how humans and VSR model learn to recognize the visemes and whether there is a critical period. 

We train two types of VSR models, i.e., monolingual and bilingual. The monolingual model is exposed to only one language (either English or Mandarin). In contrast, the bilingual model is trained in both English and Mandarin. We try to evaluate the accuracy of common visemes between the two languages in monolingual and bilingual testing environments. Also, infants show experimental influences on speech perception and its neural foundation \cite{werker2018perceptual}. We incrementally expose the VSR model with English and Mandarin data to simulate the infant's exposure to the outside world and thus the language acquisition. The initial model is trained with 25\% of the whole data, and subsequent models are trained with 50\%, 75\%, and 100\% of the dataset. 

We train three models with different dataset splits, namely, monolingual English (trained on only English), monolingual Mandarin (trained on only Mandarin), and a bilingual English-Mandarin model (with an equal amount of data from English and Mandarin). We study effects of exposure to the data and the change in occurence of the critical period during training of the model. 

\section{Experiments}

We use a state-of-the-art audio-visual speech recognition model \cite{petridis2018audio} and modify it to predict visemes given a video as an input. We perform our experiments on the largest publicly-available lipreading datasets namely Lip Reading in the Wild (LRW)\cite{Chung16} and LRW1000\cite{yang2019lrw} in English and Mandarin respectively. The videos in both datasets are very short in length, which helps in analyzing how articulation varies in a single word.
 
LRW consists of short segments (1.16 seconds) of BBC News anchors articulating 500 short words. There are more than 1000 speakers and a significant variation in background, speaker types, head pose \textit{etc}. LRW1000 is also a challenging dataset due to its large variation in scale, resolution and background clutter. There are 1000 unique word classes (referred to as `pinyins') from over 2000 speakers.

\begin{figure}
    \centering
    \includegraphics{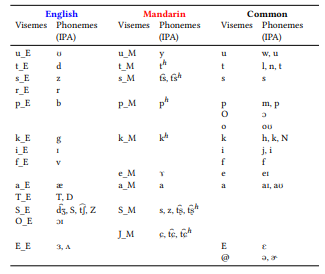}
    \caption{Viseme Classification, All symbols taken from http://nidaba.co.uk/Contents/Segments}
    \label{fig:viseme classification}
\end{figure}

Both datasets have been provided with training and testing splits. We split the training data into four equal halves by dividing the number of samples per word/pinyin. Furthermore, we train monolingual (English or Mandarin) and bilingual models (Both Engish and Mandarin) under various training splits of 25\%, 50\%, 75\%, and 100\%, and we study the impact of the amount of data used for training the models on the existence of the critical period. The datasets contain videos labeled with the corresponding word. We convert the words into visemes to train our models. The translation process is as follows:

\begin{itemize}
    \item We first use python library \textit{epitran}\footnote{http://github.com/dmort27/epitran} as a tool to transliterate text in both English and Mandarin, to International Phonetic Alphabet (IPA) format. 
    \item Secondly, we use speech marks \textit{metadeta} provided in the documentation by Amazon Polly\footnote{https://docs.aws.amazon.com/polly/latest/dg/using-speechmarks1.html}, a cloud based service that converts text to lifelike speech. Currently, it lists phonemes and the corresponding visemes for around 30 languages. We follow the tables for British English\footnote{https://docs.aws.amazon.com/polly/latest/dg/ph-table-english-uk.html} and Mandarin Chinese\footnote{https://docs.aws.amazon.com/polly/latest/dg/ph-table-mandarin.html}.
    \item As demonstrated in Table~\ref{fig:viseme classification}, we segregate visemes into three categories and use the following nomenclature. For each viseme `v':
    \begin{itemize}
        \item If `v' is common for both English and Mandarin, labelled as simply `v'.
        \item If `v' is unique for English, labelled as `v\_E'.
        \item If `v' is unique for Mandarin, labelled as `v\_M'.
    \end{itemize}
\end{itemize}

\begin{figure}[!h]
    \centering
    {%
    \includegraphics[width=0.46\linewidth]{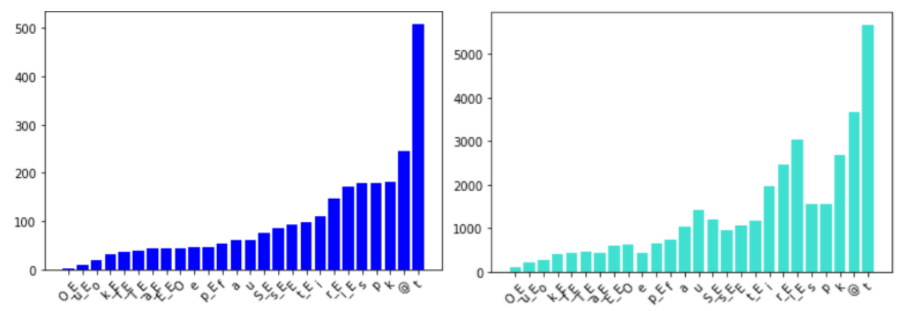}}\quad
{%
    \includegraphics[width=0.46\linewidth]{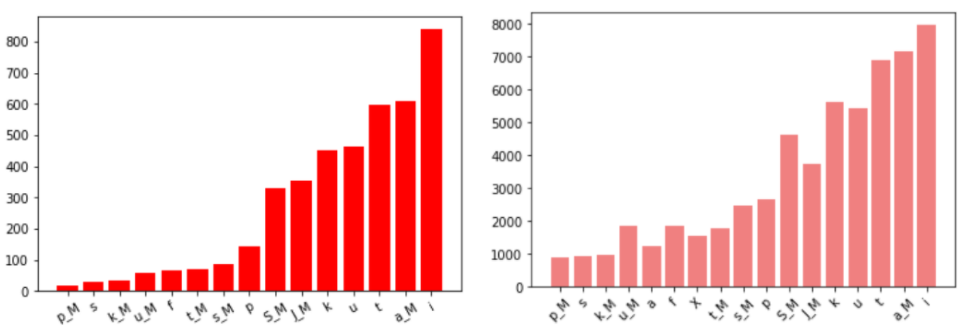}}
    \caption{Viseme Distribution for LRW, LRS2, LRW1000, CMLR \textit{(Left to Right)}}
    \label{fig:LRW and LRS2}
\end{figure}


The viseme distributions of LRW and LRW-1000 are shown in Figure~\ref{fig:LRW and LRS2}. To establish whether this distribution is similar to everyday spoken speech, we also plot the visemic distribution from the testing sets of two sentence level datasets, namely LRS2 \cite{Afouras18c} for English and CMLR \cite{zhao2019cascade} for Mandarin. We conclude that in both languages, common visemes occur with a higher frequency.

We train two types of models, i.e., monolingual and bilingual, over different training splits as explained above. Input to monolingual and bilingual models differs in the type of data used and the number of visemes to predict. 
Table ~\ref{table:viseme classification} shows the visemes specific to English and Mandarin, and it also mentions visemes common across the two languages. For monolingual models, we predict language-specific visemes and common visemes for a given video. Moreover, for bilingual models, we predict all visemes across the two languages.

We train multiple monolingual and bilingual models with an increasing amount of data, and we measure viseme classification accuracy at each epoch on the test set to check for a sudden rise in accuracy. This indicates the presence of a critical period. 

\subsection{Data Preprocessing:} We shuffle the order of the input videos at each epoch, resize them to 96×96, and then random crop them to 88×88 as the final input to the model. We select a batch of videos in each training iteration. Each video is flipped horizontally with probability 0.5, converted
to grayscale, and normalized to [0, 1]. On LRW-1000, we chose 40 frames for each word and put the target word at the center to make it similar to the data in LRW. It is found by \cite{feng2020learn} that it can provide more context and improve the performance.

\section{Results}
In this section, we demonstrate our results and provide a detailed analysis to support our hypothesis. First, we present the viseme classification accuracy over each dataset split for both monolingual scenarios and bilingual in Table~\ref{table:viseme classification accuracy}. We observe on par accuracy for each dataset split. To summarize our results, we divide this section into three categories. First, we examine whether there occurs a critical period as seen in infants during language acquisition. Secondly, we explore how these critical periods affect the above language acquisition. Finally, we compare cognitive precursors and acquisition of first and second language learners, by inspecting performance on visual speech recognition models. 
\begin{table}[!ht]
\caption{Viseme Classification Accuracy over training dataset splits} 
\label{table:viseme classification accuracy}
\begin{tabular}{ ccccc } 
\toprule
\multicolumn{1}{c}{} & \multicolumn{4}{c}{\% of Training data used} \\
{Model} & 25\% & 50\% & 75\% & 100\% \\
\midrule
Monolingual English & 83.81\% & 87.57\% & 89.88\% & 94.60\% \\ 
Monolingual Mandarin & 79.96\% & 81.22\% & 85.18\% & 87.86\% \\ 
Bilingual English-Mandarin & 82.14\% & 85.92\% & 86.31\% & 91.67\%  \\
\bottomrule
\end{tabular}
\end{table}

\begin{figure*}[!ht]
  \centering
  {%
    \includegraphics[width=0.23\linewidth]{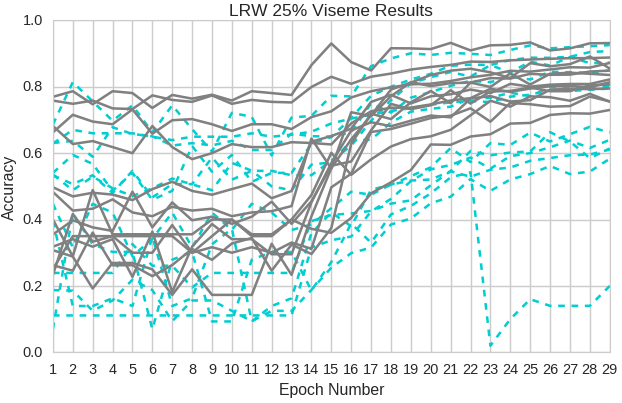}}\quad
{%
    \includegraphics[width=0.23\linewidth]{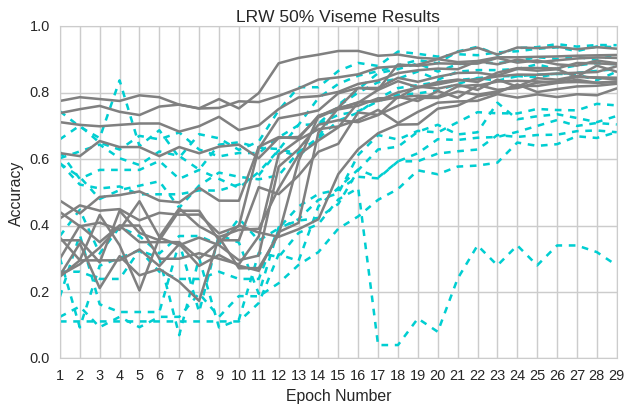}}\quad
{%
    \includegraphics[width=0.23\linewidth]{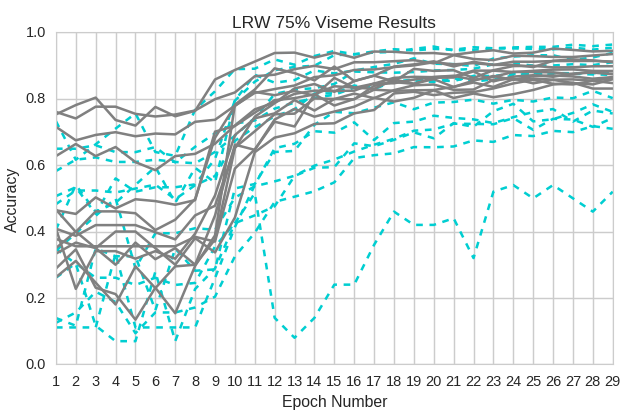}}\quad
{%
    \includegraphics[width=0.23\linewidth]{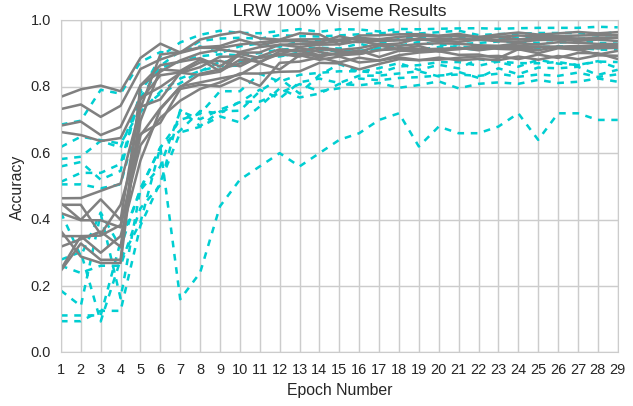}}
  \caption{Viseme Classification Accuracy for Monolingual English model on LRW dataset using 25\%(a), 50\%(b), 75\%(c), 100\%(d) training splits, respectively (left to right) across each epoch number. \textcolor{black}{\textsc{Grey}} depicts common visemes in English and Mandarin and \textcolor{blue}{\textsc{Blue}} highlights other English visemes.}
  \label{MonoEnglishAllResults}
\end{figure*}
\begin{figure*}[!ht]
  \centering
  {%
    \includegraphics[width=0.23\linewidth]{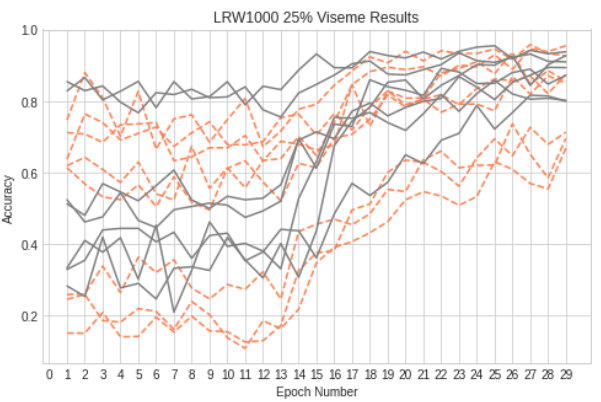}}\quad
{%
    \includegraphics[width=0.23\linewidth]{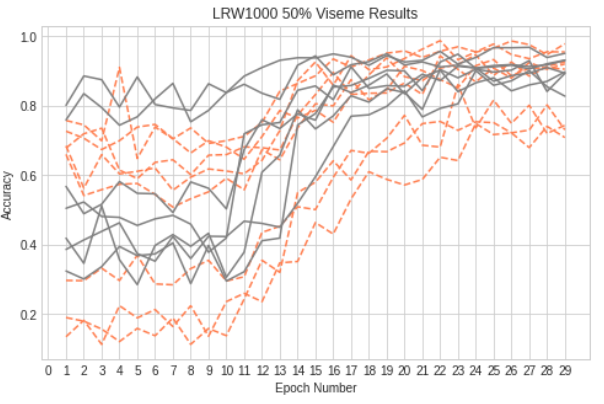}}\quad
{%
    \includegraphics[width=0.23\linewidth]{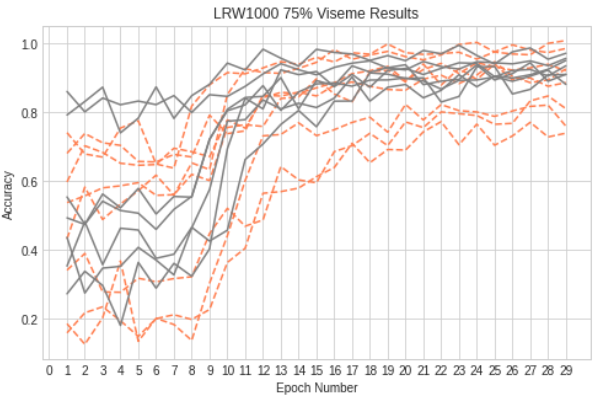}}\quad
{%
    \includegraphics[width=0.23\linewidth]{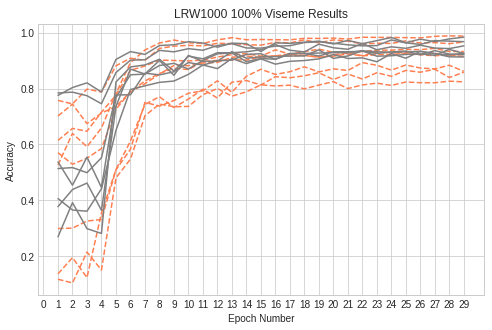}}
  \caption{Viseme Classification Accuracy for Monolingual Mandarin model on LRW1000 dataset using 25\%(a), 50\%(b), 75\%(c), 100\%(d) training splits, respectively (left to right) across each epoch number. \textcolor{black}{\textsc{Grey}} depicts common visemes in English and Mandarin and \textcolor{red}{\textsc{Red}} highlights other Mandarin visemes.}
  \label{MonoMandarinAllResults}
\end{figure*}
\begin{figure*}[!ht]
  \centering
  {%
    \includegraphics[width=0.45\linewidth]{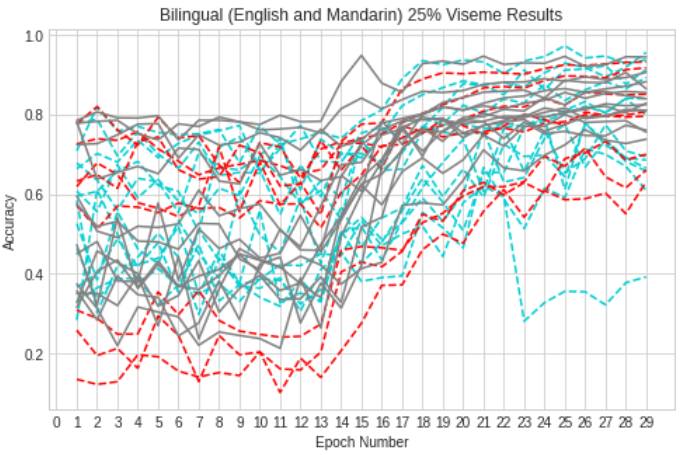}}\quad
{%
    \includegraphics[width=0.45\linewidth]{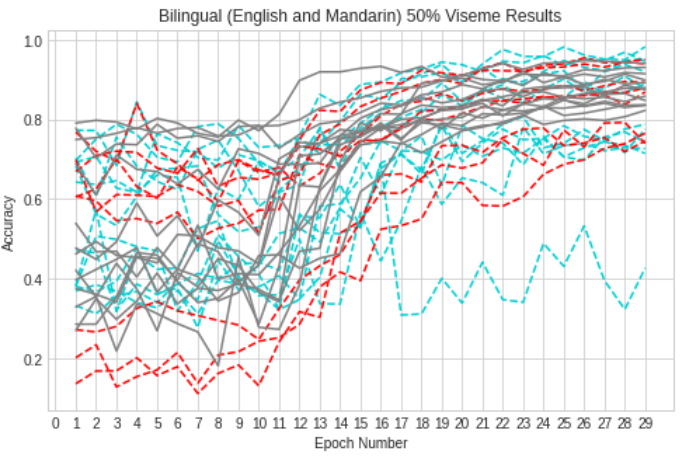}}\quad
{%
    \includegraphics[width=0.45\linewidth]{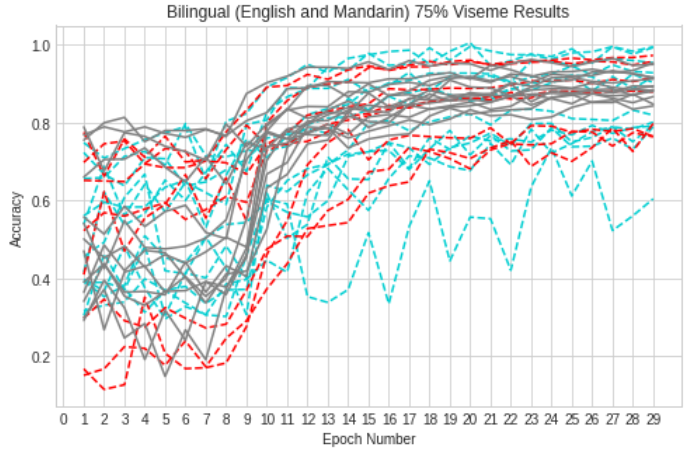}}\quad
{%
    \includegraphics[width=0.45\linewidth]{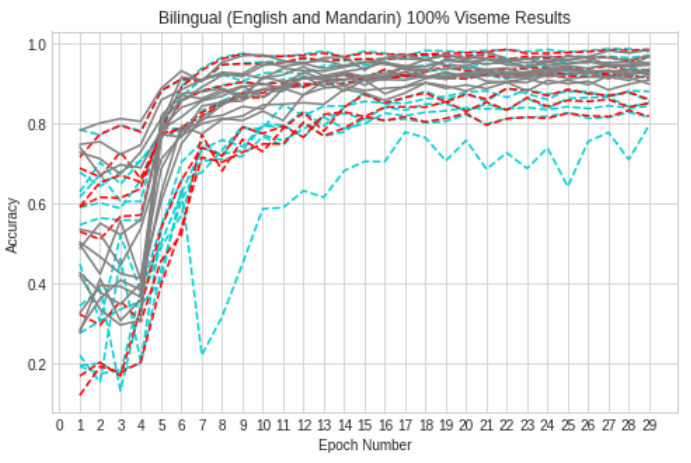}}
  \caption{Viseme Classification Accuracy for Bilingual English-Mandarin model on LRW and LRW1000 dataset using 25\%(a), 50\%(b), 75\%(c), 100\%(d) training splits, respectively (left to right) across each epoch number. \textcolor{black}{\textsc{Grey}} depicts common visemes in English and Mandarin, \textcolor{red}{\textsc{Red}} highlights other Mandarin visemes and \textcolor{blue}{\textsc{Blue}} highlights other English visemes.}
  \label{BiEnglishMandarinAllResults}
\end{figure*}
\subsection{Occurrence of Critical period} During training on monolingual and bilingual models respectively, we observe a period of surge in viseme classification accuracy on the corresponding test sets. This highlights a stage where the learning capability is maximal. This can be compared to the critical period observed by psychologists during the phase of perceptual attunement where visemes can be best distinguished. In Figure~\ref{MonoEnglishAllResults} (a), the critical period occurs around epoch number 15, as we see a soar in classification accuracy for 95\% of visemes. Moving from Figure~\ref{MonoEnglishAllResults} (b) to (d), we see a similar sensitive period for monolingual English-only models at epoch number 10 for 50\% training split, and later shifts to epoch number 7 and 5 approximately as the exposure to amount of data increases to 75\% and 100\% respectively. For monolingual Mandarin-only models and bilingual English-Mandarin models (as shown in Figure~\ref{MonoMandarinAllResults} and ~\ref{BiEnglishMandarinAllResults}), identical critical trajectories are recognised around the same time across the splits. This clearly shows how deep learning based computational models follow the same theory of language acquisition as mentioned in \cite{albareda2011acquisition} for both bilingual and monolingual environments. 

\subsection{Effects of critical periods on language acquisition}
\begin{figure}
  \centering
  {%
    \includegraphics[width=\textwidth]{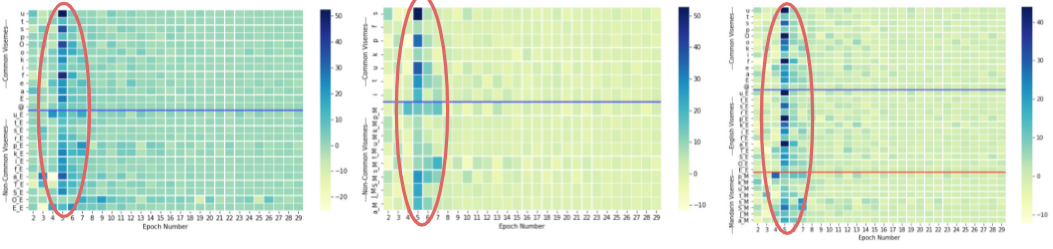}}  \caption{Heatmap representation of Viseme Classification Accuracy for Monolingual English model, Monolingual Mandarin model, Bilingual English-Mandarin model \textit{(Left to Right)}}
  \label{HeatMapsAll}
\end{figure}

The occurrence of a critical period exhibits a steep increase in the learning ability of the model as visemes become more discriminable. However, it is noticed that once this critical phase has reached, the learning ability, i.e., the relative accuracy increase per epoch, decreases and becomes stable. To visualize the same, we plot the relative viseme classification accuracy using heat maps as shown in Figure ~\ref{HeatMapsAll}. The relative increase in accuracy from the previous epoch is highest during the critical period \textit{(circled in red for each case)} and then it starts to decline and stabilizes after a few epochs. Furthermore, the critical period occurs later if the model is exposed to less data, as shown in Figure ~\ref{MonoEnglishAllResults} and ~\ref{MonoMandarinAllResults}. Consequently,  language acquisition depends on the occurrence of the critical period, which is affected by the amount of exposure to the language, i.e., language acquisition would be quicker in linguistically rich environments. 

\begin{figure}[!ht]
  \centering
  {%
    \includegraphics[width=\textwidth]{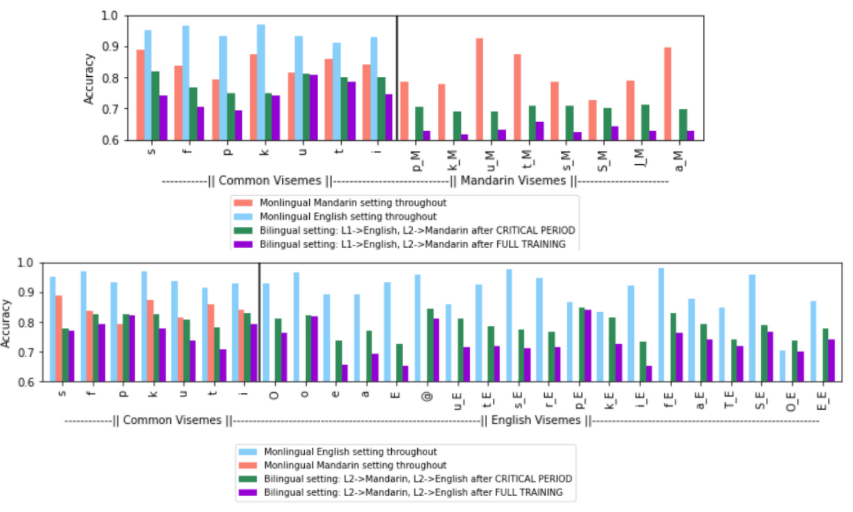}}  \caption{Accuracy for each viseme. \textcolor{red}{\textsc{Red}} and \textcolor{blue}{\textsc{Blue}} bars depict accuracies for Monolingual Mandarin and English training throughout. \textcolor{green}{\textsc{Green}} denotes L2 performance after critical period of L1 and \textcolor{purple}{\textsc{Purple}} denotes L2 performance after full training of L1.}
  \label{crossInferenceResults}
\end{figure}

\subsection{Impact of L1 and L2 learning performance in Bilingual setting}
According to the critical period hypothesis, there is an ideal time window to acquire language in a linguistically rich environment, after which further language acquisition becomes more difficult and effortful. It also states that the first few years of life are crucial when an individual can acquire a language if presented with adequate stimuli and that language acquisition relies on neuroplasticity. Literature mentions that if language input does not occur until after this time, the individual will never achieve a complete command of the language \cite{Nielsen1960SpeechAB, lenneberg1967}. 
We inspect this theory by experimenting using our visual speech recognition models. We simulate language acquisition in a bilingual environment for English and Mandarin languages and compare them to monolingual English and Mandarin models. The models are trained in two ways; in the first case, the model is trained on L1 language till the \textsc{critical period} is achieved, and then it is switched to L2 second language and trained till the model converges. In the other scenario, the model is trained on L1 language \textsc{till convergence} and then switched to L2 second language and trained till convergence. In both cases, the model inferences on L2 second language. We express the accuracy of each viseme in Figure ~\ref{crossInferenceResults}. \textsc{Green} and \textsc{Purple} bars highlight the cross inference results switched after the critical period and full training, respectively.

Between the training schemes discussed above, we see an average drop of 6\% in accuracy on common visemes, while 11\% and 13\% drop in accuracy for non-common visemes of Mandarin and English respectively. These results support our claim and show that till the \textsc{critical period}, the model can be compared to a child that has the ability to analyze sounds and build sets of rules for morphology and syntax from the spoken language around them. While training till convergence, the model can be compared to an adult whose abstract phonological grasping ability gradually declines. Comparing with monolingual training throughout, we see a higher accuracy gain, reinforcing that humans learn L1 as a natural process, without any active intervention. 


\section{Conclusion}
Through our experiments, we conclude that we can draw several connections between theories in cognitive psychology related to early language acquisition and methodological computational modelling. We underline the presence of a critical learning period and emphasize on the effects of the same. We also briefly analogize theories in second language acquisition and the importance of critical period during this learning adaption.  

\bibliographystyle{ACM-Reference-Format}
\bibliography{sample-manuscript}
\end{document}